\documentclass[conference]{IEEEtran}
\IEEEoverridecommandlockouts
\usepackage{cite}
\usepackage{amsmath,amssymb,amsfonts}
\usepackage{algorithmic}
\usepackage{graphicx}
\usepackage{textcomp}
\usepackage{xcolor}

\usepackage{todonotes}
\usepackage[caption=false]{subfig}
\usepackage[export]{adjustbox}
\usepackage{hyperref}
\usepackage{cite}

\newcommand\Tstrut{\rule{0pt}{2.6ex}}  

\usepackage{tikz}
\usepackage{textcomp}
\usepackage{hyperref}
\usepackage{lipsum}

\newcommand\copyrighttext{%
  \footnotesize \textcopyright 2022 IEEE. Personal use of this material is permitted.
  Permission from IEEE must be obtained for all other uses, in any current or future
  media, including reprinting/republishing this material for advertising or promotional
  purposes, creating new collective works, for resale or redistribution to servers or
  lists, or reuse of any copyrighted component of this work in other works.
  DOI: \href{https://doi.org/10.1109/IJCNN55064.2022.9892562}{10.1109/IJCNN55064.2022.9892562}}
\newcommand\copyrightnotice{%
\begin{tikzpicture}[remember picture,overlay]
\node[anchor=south,yshift=10pt] at (current page.south) {\fbox{\parbox{\dimexpr\textwidth-\fboxsep-\fboxrule\relax}{\copyrighttext}}};
\end{tikzpicture}%
}

\def\BibTeX{{\rm B\kern-.05em{\sc i\kern-.025em b}\kern-.08em
    T\kern-.1667em\lower.7ex\hbox{E}\kern-.125emX}}

\begin{document}

\definecolor{mygreen}{rgb}{0.01, 0.75, 0.24}


\title{Graph-Based Multi-Camera Soccer Player Tracker}

\author{
\IEEEauthorblockN{Jacek Komorowski}
\IEEEauthorblockA{\textit{Warsaw University of Technology} \\
\textit{Sport Algorithmics and Gaming}\\
Warsaw, Poland \\
jacek.komorowski@pw.edu.pl}
\and
\IEEEauthorblockN{Grzegorz Kurzejamski}
\IEEEauthorblockA{\textit{Sport Algorithmics and Gaming} \\
Warsaw, Poland \\
g.kurzejamski@sagsport.com}
}


\maketitle

\copyrightnotice

\begin{abstract}
   The paper presents a multi-camera tracking method intended for tracking soccer players in long shot video recordings from multiple calibrated cameras installed around the playing field.
   The large distance to the camera makes it difficult to visually distinguish individual players, which adversely affects the performance of traditional solutions relying on the appearance of tracked objects. 
   Our method focuses on individual player dynamics and interactions between neighborhood players to improve tracking performance.
   To overcome the difficulty of reliably merging detections from multiple cameras in the presence of calibration errors, we propose the novel tracking approach, where the tracker operates directly on raw detection heat maps from multiple cameras.
  Our model is trained on a large synthetic dataset generated using Google Research Football Environment and fine-tuned using real-world data to reduce costs involved with ground truth preparation.
\end{abstract}




\section{Introduction}


Reliable tracking of players is a critical component of any automated video-based soccer analytics solution~\cite{bornn2018soccer, stensland2014bagadus}.
Player tracks extracted from video recordings are the basis for players' performance data collection, events tagging, and tactical and player fitness information generation.
This paper presents a multi-camera multi-object tracking method suitable for tracking soccer players in long-shot videos. 
Our method is intended for tracking multiple players from a few cameras (e.g. 4-6) installed at fixed positions around the playing field. An exemplary input from such a multi-camera system is shown in Fig~\ref{fig:gt2_4cameras}. 



\begin{figure}[t]
\begin{center}
\includegraphics[width=1.0\linewidth, trim={0 1cm 0 2.2cm},clip]{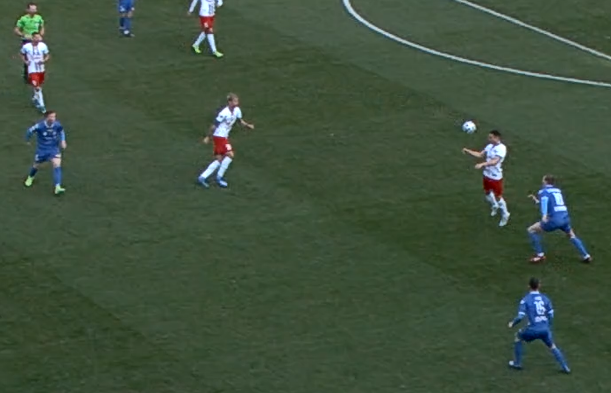}
\end{center}
\caption{A cropped region from a long shot camera view illustrating the difficulty of player identification using visual features. Players from the same team have a similar appearance and jersey numbers are mostly not recognizable.}
\label{fig:closeup1}
\end{figure}

Such a setup makes tracking players a very challenging problem.
Most of the existing multi-object tracking methods~\cite{bredereck2012data, zhang2015online, xu2016multi, wen2017multi} follow \emph{tracking-by-detection} paradigm and heavily rely on the appearance of tracked objects to link detections between consecutive video frames into individual tracks.
In our case scenario, it's very difficult to distinguish players based on their appearance.
Players from the same team wear jerseys of the same color.
Cameras are distant from the ground plane and cover a large part of the pitch. Thus images of players are relatively small. As seen in Fig.~,\ref{fig:closeup1} players' jersey numbers and facial features are hardly visible. Frequent occlusions, when players rush towards the ball, make the problem even more challenging.
Another problem is how to reliably aggregate player detections from multiple cameras in the presence of camera calibration errors, inaccuracies in single-camera player detections and player occlusions.



\begin{figure}
\begin{center}
\includegraphics[width=1.0\linewidth, trim={0 2cm 14cm 2cm},clip]{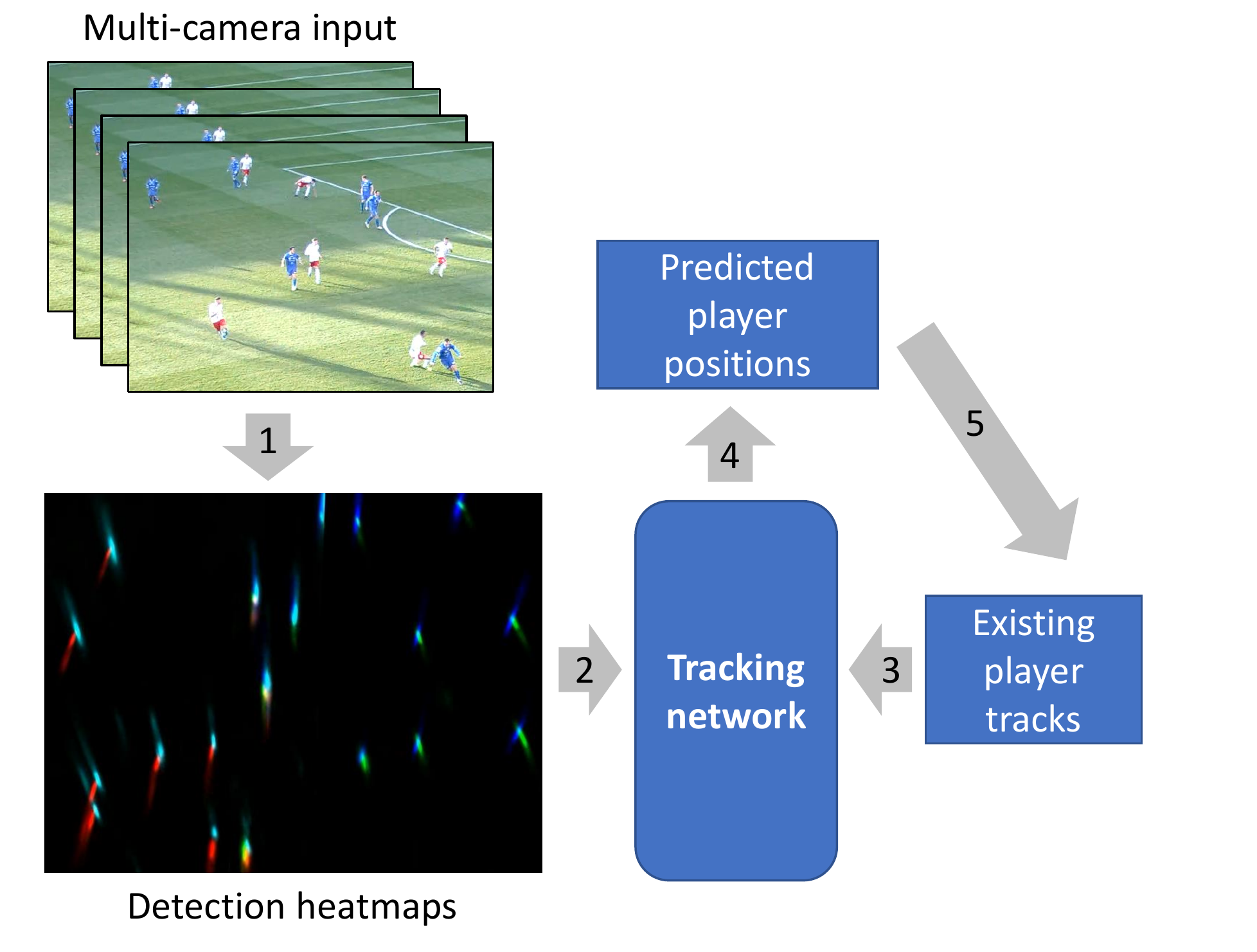}
\end{center}
\caption{Overview of our tracking solution. Detection heat maps from single-camera detectors are transformed into bird's-eye view using homographies and stacked as a multi-dimensional tensor, with one dimension per camera. For visualization purposes, we color-code detection maps from different cameras (1). The tracking network uses detection heat maps (2) and existing player tracks (3) to regress new player positions (4). Finally, existing tracks are extended using regressed player positions. A simple heuristic is used to terminate tracks or initiate new tracks (5).}
\label{fig:overview}
\end{figure}

This paper presents a multi-camera multi-object tracking method intended for tracking soccer players in long shot video recordings from cameras installed around the playing field. 
We assume cameras are initially calibrated and homographies between a camera plane and a ground plane are estimated by manually matching distinctive points (e.g. four corners of the playing field) in each camera view with the bird's-eye view diagram of the pitch. 
Our solution is an online tracking method, where time-synchronized frames from cameras are processed sequentially and detections at each timestep are used to extend existing or initiate new tracks.
In the prevailing tracking-by-detection paradigm~\cite{bredereck2012data}, detections at each timestep are linked with existing players' tracks by solving the so-called \emph{assignment problem}. 
The problem is not trivial to solve in a single-camera setup but becomes much more complicated for multiple-camera installations. 
Detections of the same player from multiple cameras must be first aggregated before linking them with existing tracks. This is a challenging task due to occlusions, camera calibration errors, and inaccuracies in single-camera detector outputs.
Recent methods use sophisticated techniques with explicit occlusion modeling, such as Probabilistic Occupancy Maps~\cite{fleuret2007multicamera, zhang2020multi, liang2020multi} or generative models consisting of Convolutional Neural Network and Conditional Random Fields~\cite{baque2017deep}, to reason about objects ground plane positions consistent with detections from multiple cameras. Such aggregated bird's-eye view detection map is fed into the tracker.
On the contrary, our method follows \emph{tracking-by-regression} principle and does not need such preprocessing step. 
It processes raw detection maps from each camera before the non-maxima suppression step. 
Aggregation of multiple single-camera detector outputs is moved from the preprocessing stage into the tracking method itself and is end-to-end learnable.
This is achieved by using a homography to transform detection heat maps from multiple cameras onto the common bird's-eye view plane and stacking them as a multi-channel tensor. Each channel corresponds to one camera view. 
Extracting and aggregating information from multiple detection maps is done within the tracking network itself.
In the absence of discriminative players' appearance cues, our method focuses on individual player dynamics and interactions between neighborhood players to improve the tracking performance and reduce the number of identity switches.
We model player dynamics using an LSTM-based Recurrent Neural Networks (RNN)~\cite{gers2002learning} and interaction between players using a Graph Neural Network (GNN) with message passing mechanism~\cite{gilmer2017neural}.

\begin{figure}[t]
\begin{center}
\includegraphics[width=1.0\linewidth, trim={2cm 2cm 5cm 3cm},clip]{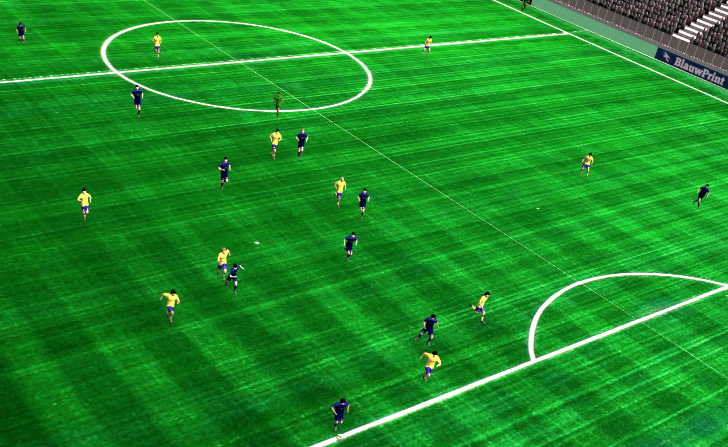}
\end{center}
\caption{A crop from a synthetic video sequence generated using Google Research Football Environment.}
\label{fig:gfootball1}
\end{figure}

Our learnable tracking method, using deep networks to fuse detections from multiple cameras and model player dynamics/interactions, requires a large amount of annotated data for training. 
Manual labeling of player tracks in a sufficiently large number of video recordings is prohibitively expensive.
We train our model using the synthetic data generated with Google Research Football Environment~\cite{kurach2020google} (GRF) to overcome this problem. We adapted GRF code to allow recording time-synchronized videos from four virtual cameras placed at the fixed locations around the playing field with accompanying ground truth player tracks.
An exemplary frame from a soccer game video generated using GRF is shown in Fig.~\ref{fig:gfootball1}.
To bridge the domain gap between real and synthetic data, we employ a two-step training approach. First, our model is trained on a large synthetic dataset generated using GRF environment, and then it's fine-tuned using a smaller, manually labeled real-world dataset.

In summary main contributions of our work are as follows. First, we present a learning-based approach for multi-camera multi-object tracking, where aggregation of multiple single-camera detector outputs is a part of the end-to-end learnable tracking network and does not need a sophisticated preprocessing.
Second, our tracking solution exploits player dynamics and interactions, allowing efficient tracking from long shot video recordings, where it's difficult to distinguish individual objects based on their visual appearance.


\section{Related work}

\textbf{Multi-object tracking}. 
The majority of recent multi-object tracking methods follow the 
\emph{tracking-by-detection} paradigm, which splits the problem into two separate phases. First, objects are detected in each frame using a pretrained object detector, such as YOLO~\cite{redmon2016you}. 
Then, detections from consecutive frames, usually in the form of object bounding boxes, are linked together to form individual tracks.
These methods can be split into two categories: online and offline approaches.
Online methods~\cite{milan2017online, chu2017online} sequentially process incoming video frames. Detections from a new frame are compared with existing tracks and either linked with existing tracks or used to initiate a new track (so-called \emph{data association} problem). The similarity between detection and an exiting track is computed, based on positional, visual appearance, or motion cues, using different techniques, such as Recurrent Neural Networks~\cite{milan2017online} or attention mechanism~\cite{chu2017online}.
Online approaches use global optimization techniques, such as graph optimization~\cite{zhang2008global, shitrit2013multi} or hierarchical methods~\cite{ristani2018features}, to find optimal tracks over large batches of frames or even entire video sequences.

\textbf{Multi-camera tracking.}
Existing Multi-Camera Multi-Object (MCMO) tracking methods can be split into two main groups.
Earlier methods track targets separately in each view, then merge tracklets recovered from multiple views to maintain target identities~\cite{kang2004tracking, wen2017multi, he2020multi}. 
E.g.~\cite{wen2017multi} encodes constraints on 3D geometry, appearance, motion continuity, and trajectory smoothness among 2D tracklets using a space-time-view hyper-graph. Consistent 3D trajectories are recovered by finding dense sub-hypergraphs using a sampling-based approach.
Single-camera tracking methods are sensitive to occlusions and detection misses, which can lead to track fragmentation.
Fusion of fragmented tracklets to generate consistent trajectories maintaining identities of tracked targets is a challenging task.

The second group of methods first aggregates detections from multiple cameras at each time step, then links aggregated detections to form individual tracks using a single-camera tracking approach.
\cite{fleuret2007multicamera} constructs a Probability Occupancy Map (POM) to model the probability of a person's presence in discrete locations in a bird's-eye view plan of an observed scene. The occupancy map is created using a generative model by finding the most probable person locations given background-segmented observations from multiple cameras.
This concept is extended in ~\cite{zhang2020multi} by incorporating sparse players' identity information into the bird's-eye view occupancy map. Aggregated detections are linked into tracks using a graph-based optimization (k-Shortest Path).
\cite{baque2017deep} aggregates detections from multiple cameras using a convolutional neural network and Conditional Random Field (CRF) to model potential occlusions on a discretized ground plane. The method is trained end-to-end and outputs probabilities of an object's presence in each ground plane location. The aggregated detections are linked into tracks using a graph-based optimization.
\cite{lima2021generalizable} formulates detection fusion problem as a clique cover problem. Appearance features are exploited during the fusion using a person re-identification model.

\section{Multi-camera soccer players tracker}

Multi-Object Tracking (MOT) aims to detect multiple targets at each frame and match their identities in different frames, producing a set of trajectories over time.
Our setup uses video streams produced by a few (from 4 to 6) high-definition cameras installed around the playing pitch. See Fig.~\ref{fig:gt2_4cameras} for an exemplary input.
We assume cameras are initially calibrated. A homography between each camera plane and a ground plane is estimated by manually matching distinctive points (e.g. four corners of the playing field) in each camera view with the bird's-eye view diagram of the pitch. 

A popular \emph{tracking-by-detection} paradigm requires prior aggregation of detections from multiple cameras at each time step. In our camera configuration, this is problematic due to the following factors.
The initial camera calibrations become less accurate over time due to environmental factors, such as strong wind or temperature variations altering the length of metal elements to which cameras are fixed.
A software time synchronization mechanism is not perfect, and due to the network jitter, there  are inaccuracies in timestamps embedded in recorded video frames.
There are frequent occlusions as players compete for the ball.
As a result, it's difficult to reliably link detections from multiple cameras corresponding to the same player.

Instead of developing and fine-tuning a heuristic for aggregating detections of the same player from different cameras, we choose a learning-based approach based on raw outputs from multiple single-camera detectors.
Contrary to the typical approach, we do not use players' bounding boxes detected in each camera view.
We take raw detection heat maps from a pretrained player's feet detector as an input.
For this purpose, we use FootAndBall~\cite{footandball} detector, modified and trained to detect a single class: player's feet (to be more precise, the center point between two players' feet).
In the detector, we omit the last non-maxima suppression (NMS) and bounding box calculation steps and output raw detection heat maps for player feet class. A detection heat map is a single channel tensor whose values can be interpreted as the likehood of a player's feet presence at a given location. 

The idea behind our method is illustrated in Fig.~\ref{fig:overview}.
Detection heat maps from multiple single-camera detectors at a time step $t_k$ are transformed into bird's-eye view using homographies and stacked as a multi-dimensional tensor, with one dimension per camera (1). Note that detections of the same player from different cameras, shown in different colors, sometimes have little or no overlap.
The \textbf{tracking network} uses detection heat maps (2) and existing player tracks (3) to regress new player positions at the time step $t_k$ (4).
Existing tracks are extended by appended the regressed player positions (5). 
We use a simple heuristic to initiate new tracks or terminate inactive tracks. 
Details of each component are given in the following sections.

\subsection{Tracking network architecture}



\begin{figure*}[t]
\begin{center}
\includegraphics[width=0.9\linewidth, trim={0.5cm 7.4cm 0.5cm 0.2cm},clip]{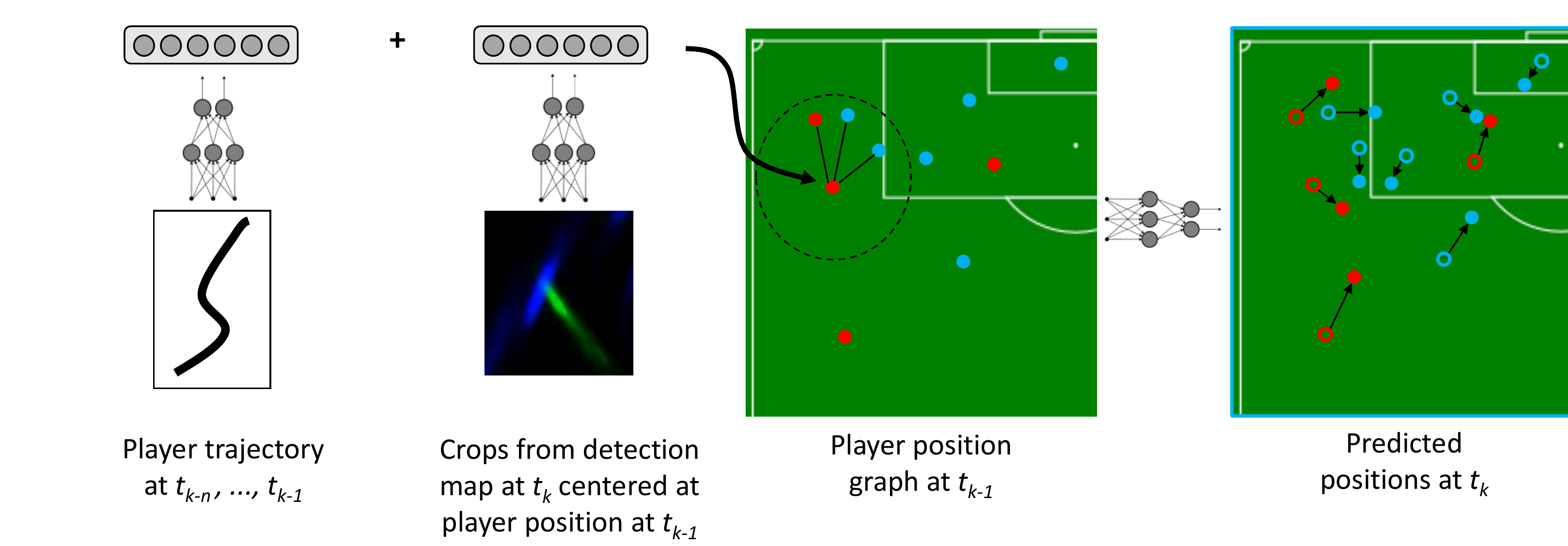}
\end{center}
\caption{High-level architecture of our tracking network. 
For each player, we encode its previous trajectory and stacked crops from detections maps at time step $t_k$ centered at the last player position at $t_{k-1}$. Position graph is built, where each player corresponds to a graph node and edges connect neighborhood players (for clarity, only edges from one player are shown). 
Vertex attributes are formed by concatenating trajectory and detection embeddings. Edge attributes are initialized with a relative position between players.
Information between neighborhood nodes is exchanged using a message passing algorithm, and vertex attributes are updated.
Finally, a player position at time $t_k$ is regressed based on updated vertex attributes.
}
\label{fig:high_level}
\end{figure*}

The aim of the \textbf{tracking network} is to regress the new position of tracked players (at a time step $t_k$), based on their previous trajectory (up to a time step $t_{k-1}$), interaction with other players, and output from player feet detectors (at a time step $t_k$). 
For this purpose, we use a Graph Neural Network~\cite{gilmer2017neural}, where each node encodes player state and interaction between neighborhood players is modeled using a message passing mechanism.
The high-level overview of the tracking network architecture is shown in Fig.~\ref{fig:high_level}.
As an input, we use detection heat maps from player's feet detectors transformed into a bird's-eye view using homographies and stacked together to form a multi-channel tensor $\mathcal{T}_D$, with each channel corresponding to one camera.
For computational efficiency, for each tracked player, we take a rectangular crop from $\mathcal{T}_D$, centered at the last known player position at a time step $t_{k-1}$. The left side of Fig.~\ref{fig:high_level} shows a color-coded example of such a crop, with each color corresponding to one channel (one camera).
For each player, we use a feed-forward neural network to compute an embedding of its corresponding crop from the detection map $\mathcal{T}_D$ and a recurrent neural network for the embedding of its previous track. We concatenate these two embeddings to form a fused player embedding $\textbf{p}_i$.
We model interactions between neighborhood players' by building an undirected graph.
Each node represents a player, and its initial state is set to a fused player embedding $\textbf{p}_i$.
All players in the radius of $k=3$ meters are connected with edges, and the distance between players (computed using their last known positions in a time step $t_{k-1}$) is used as an initial edge attribute.
See the middle part of Fig.~\ref{fig:high_level} for visualization of the positions graph. Note that for simplicity, only edges originating from one node (player) are shown.
Information between neighborhood nodes in the graph is exchanged using a message passing algorithm~\cite{gilmer2017neural}, and node attributes are updated.
Finally, a new position of each player at a time step $t_k$ is regressed using its corresponding node attributes.
A player's track is extended using this regressed position.


\paragraph{Detections encoding}. For each tracked player, we take a rectangular crop from stacked bird's-eye view detection maps $\mathcal{T}_D$ centered at the last known player position at a time step $t_{k-1}$. In our implementation, each crop is a 4x81x81 tensor, with the number of channels equal to the number of cameras and 81x81 spatial dimensions.
Multiple detection maps are aggregated using a convolutional layer with a 1x1 kernel, producing a one-dimensional feature map.
Then, the crop is downsampled to 32x32 size, flattened, and processed using a detection encoding network $f_D$, a multi-layer perceptron (MLP) with four layers having 1024, 512, 256, and 128 neurons and ReLU non-linearity. 
This produces 128-dimensional detection embedding.
Detection encoding subnetwork is intended to extract information about probable positions of the player's feet at a time step $t_k$, taking into account detection results from all cameras.
We experimented with other architectures, such as 2D convolutions, but they produced worse results. See the ablation study section for more details.



\paragraph{Trajectory encoding} uses a simple recurrent neural network with a single-layer LSTM cell.
For each tracked player, at a time step $t_k$ it accepts its previous trajectory $(x_{k-n}, y_{k-n}), \ldots, (x_{k-1}, y_{k-1})$ and produces 128-dimensional trajectory embedding.
It should be noted that there's no need to process the entire player trajectory at each time step at the inference stage.
For each tracked player, at a time step $t_k$ , we feed only previous player position $(x_{k-1}, y_{k-1})$ to trajectory encoding subnetwork and use the LSTM state of the prior step.

\paragraph{Position graph} models interactions between neighborhood players.
We build an undirected graph, where each node represents a tracked player.
For each player, we compute a 256-dimensional, fused player embedding $\mathbf{p}_i$ by concatenating its detection and trajectory embeddings.
Initial features $\mathbf{x}_i^{(0)}$ of $i$-th graph vertex are set to this concatenated player embedding $\mathbf{p}_i$.
To model interactions between neighborhood players, we add an edge between all pairs of vertices that are closer than $3$ meter threshold.
Edge features $\mathbf{e}_{i,j}$ of an $(i, j)$ edge are set as the relative position of the $j$-th player with respect to the $i$-th player at the previous time step $t_{k-1}$.  
$\mathbf{e}_{i,j} = (x^{j}_{k-1} - x^{i}_{k-1}, y^{j}_{k-1} - y^{i}_{k-1})$, where 
$x^{i}_{k-1}, y^{i}_{k-1}$ are coordinates of $i$-th player in the world reference frame at the time step $t_{k-1}$.
See middle part of Fig.~\ref{fig:high_level} for visualization of the positions graph. For readability only edges originating from one player are shown.

To model interaction between neighborhood players, we run two iterations of a message passing algorithm~\cite{gilmer2017neural}.
First, messages between pairs of neighbourhood nodes in iteration $r$, where $1 \leq r \leq 2$, are calculated using the following equation:
\begin{equation}
    \mathbf{m}_{i,j}^{(r)} =
    f_M
    \left(
    \mathbf{x}_i^{(r-1)},
    \mathbf{x}_j^{(r-1)},
    \mathbf{e}_{i,j}
    \right) \\,
\end{equation}
where $f_M$ is a message generation function modeled using a neural network.
$f_M$ takes features of two neighbourhood nodes $\mathbf{x}_i^{(r-1)}$, $\mathbf{x}_j^{(r-1)}$ and edge features $\mathbf{e}_{i,j}$, concatenates them and passes through a two layer MLP with 128 and 32 neurons and ReLU non-linearity. It produces a 32-dimensional message $\mathbf{m}_{i,j}^{(r)}$.

Then, messages from neighbourhood nodes are aggregated and used to update node features, using the below equation: 
\begin{equation}
    \mathbf{x}_i^{(r)} = f_N
    \left( 
    \mathbf{x}_i^{(r-1)}, 
    \frac{1}{|N(i)|}
    \sum_{j \in N(i)}
    \mathbf{m}_{i,j}^{(r)}
    \right) \\ ,
\end{equation}
where $f_N$ is a node update function.
$f_N$ concatenates node features $\mathbf{x}_i^{(k-1)}$ from the previous message passing iteration with aggregated messages from neighborhood nodes and passes through a two-layer MLP with 384 and 384 neurons and ReLU non-linearity. It produces updated node features $\mathbf{x}_i^{(k)}$.

\paragraph{Regression of a players' position} is the final element of the processing pipeline. 
The node features $\mathbf{x}_i^{(n)}$ obtained after two rounds of message passing are fed to the neural network $f_P$, which regresses $i$-th player position $(x_i, y_i)$ in a world coordinate frame at a time step $t_k$.

Details of message generation network $f_M$, node update network $f_N$, and a position regressor $f_P$ are given in Table~\ref{jk:tab:details}.

\begin{table}
\caption{Details of the network architecture. MLP is a multi-layer perceptron with a number of neurons in each layer given in brackets and ReLU non-linearity.}
\begin{center}
\begin{tabular}{l@{\quad}l@{\quad}l}
\begin{tabular}{@{}c@{}}Subnetwork \end{tabular}
& \begin{tabular}{@{}c@{}}Function \end{tabular}
& \begin{tabular}{@{}c@{}}Details \end{tabular}
\\
[2pt]
\hline
\Tstrut
$f_D$ & detection embedding net & Conv (1x1 filter, 1 out channel) \\
 & & MLP (1024, 512, 256, 128) \\
$f_T$ & trajectory encoding & single-layer LSTM \\
$f_M$ & message generation & MLP (128, 32) \\
$f_N$ & node update & MLP (384, 384) \\
$f_P$ & position regressor & MLP (128, 2) \\
[2pt]
\hline
\end{tabular}
\end{center}
\label{jk:tab:details}
\end{table}

\subsection{Network training}

Out tracker is trained end-to-end, taking as an input a bird's-eye view player detection map at a time step $t_k$; a sequence of previous raw video frames from each camera at time steps $t_{k-n}, \ldots, t_k$; and previous positions of tracked players at time steps $t_{k-n-1}, \ldots, t_{k}$.
At first, we crop rectangular regions around each player from both bird's-eye view player detection map and a sequence of previous raw video frames from each camera. These crops are centered at the previous position of the tracked player. Crops from a bird's-eye view detection map are centered at a player position at $t_{k-1}$. Crops from previous raw video frames are centered at player position at time steps $t_{k-n-1}, \ldots, t_{k-1}$.
The reason is that we do not know what the current player position (at timestep $t_k$) is at the inference stage. We are going to regress it. 
We need to center crops at the last known position of each tracked player (at timestep $t_{k-1}$).

We should note that processing the  sequence of previous video frames is needed only during the network training. 
In the inference phase, we use RNN hidden state to carry the information from previous frames.

The network predicts the new position of each tracked player at a time step $t_k$ in a world coordinate frame.
The network is trained using a mean squared error (MSE) loss defined as:
$
\mathcal{L} = \sum_{i} 
\left (
p_i^{(t)} - \hat{p}_i^{(t)}
\right)
$, 
where $\hat{p}_i^{(t)}$ is the regressed position of $i$-th player at the timestep $t_k$ and $p_i^{(t)}$ is the ground truth position.

We train our network using synthetic videos and ground truth player tracks generated using Google Research Football environment~\cite{kurach2020google} and fine-tune using a smaller real-world dataset.
The description of datasets is given in Section~\ref{sec:dataset}.

\subsection{Track initialization and terminal}
The main focus of this work is the tracking network, intended to regress a new position of tracked players in each timestep, based on their previous motion trajectory, the interaction between neighborhood players, and input detection maps.
We use simple heuristics for the track initialization and termination,
Tracks are initialized by extracting local maxima from aggregated detection heatmaps transformed to a bird's-eye view.
A new tracked object is initiated if there's a local maximum detected at a position different from the positions of already tracked objects.
We terminate a tracked object if no local maxima are detected in its vicinity for $k=20$ consecutive frames.

\section{Experimental results}

\subsection{Datasets and evaluation methodology}
\label{sec:dataset}

\paragraph{Training dataset} 

Initially, we used manually labeled data from video recordings of live events to train our system.
However, manual labeling turned out to be a very labor-intensive and error-prone task. 
For each player, we need to tag its track on each of four cameras and preserve its identity across different cameras.
It's not a trivial task, as the jersey numbers are often not readable due to the player pose, occlusions, or large distance from the camera. In our initial experiments, training with a small set of manually labeled data resulted in poor generalization.
Instead, we resort to synthetic data generated using the Football Engine from Google Research Football environment~\cite{kurach2020google}.
The Football Engine is an advanced football simulator intended for reinforcement learning research that supports all the major football rules such as kickoffs, goals, fouls, cards, corner, and penalty kicks.
The original environment generates videos from one moving camera that focuses on the playfield part with a player possessing the ball.
We modified the code to generate videos from four views at fixed locations around the pitch. This simulates the real-world setup where four fixed-view cameras are mounted on poles around the playfield. Generated videos are split into episodes, where each episode is a part of the game with a continuous player trajectory. 
When an event resulting in players' teleportation in the game engine happens, such as a goal, a new episode is started.
Altogether, we generated 418 episodes spanning almost 0.5 million time steps, each containing videos from four virtual cameras with accompanying ground truth player tracks in a world coordinate frame.
Fig~\ref{fig:gfootball1} shows an exemplary view from our modified Football Engine setup.



\paragraph{Evaluation dataset} 

\begin{figure*}[t]
\begin{center}
\includegraphics[width=0.9\linewidth]{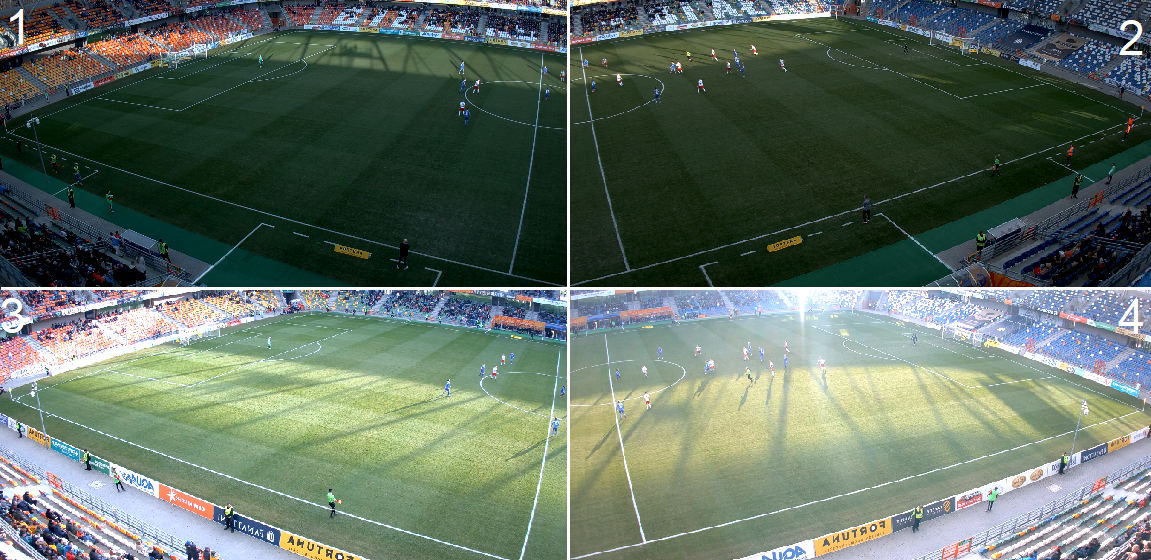}
\end{center}
\caption{Four synchronized views from the GT2 evaluation sequence. Views from cameras 3 and 4 are horizontally flipped to align with views from the other side of the pitch. Cameras on the same side (left: 1-3 and right: 2-4) show the same part of the playing field from two sides.}
\label{fig:gt2_4cameras}
\end{figure*}

Evaluation is done using a synthetic evaluation set and two real-world evaluation sets containing manually tagged recordings of real matches from two locations.
The synthetic evaluation set, denoted as GRF, consists of 4 game episodes containing recordings from four virtual cameras. 
Video rendering parameters and virtual cameras configuration are the same as in the training environment. 
To verify if the model trained on the synthetic data works on real-world data, we use two manually tagged sets of recordings, named GT2 and GT3.
Each consists of 5 minutes (12 thousand frames at 30 fps) of manually tagged recordings from four cameras installed at two different stadiums.
The camera configuration at each location is similar to the configuration we use in the Football Engine virtual environment.
Four synchronized frames from the GT2 sequence are shown in Fig.~\ref{fig:gt2_4cameras}. 
Evaluation is done using the first 2000 frames (corresponding to 100 seconds of the game) from each episode.

\paragraph{Evaluation metrics} 

We follow the same evaluation protocol as used in MOT Challenges 2019~\cite{dendorfer2019cvpr19}.
First, the correspondences between the ground truth tracks and predicted tracks are established using Kuhn-Munkres algorithm~\cite{munkres1957algorithms}.
After establishing track correspondences, we calculate MOTA, IDSW, mostly tracked, partially tracked, and mostly loss metrics.
MOTA (Multiple Object Tracking Accuracy) is a popular metric to report overall multi-object tracker accuracy, based on a number of false positives, false negatives, identity switches, and ground truth objects. 
For MOTA definition, please refer to~\cite{dendorfer2019cvpr19}.
An identity switch error (IDSW) is counted if a ground truth target $i$ is matched in the new frame to a different track than in the previous frame.
Each ground truth trajectory is classified as mostly tracked (MT), partially tracked (PT), and mostly lost (ML). 
A target successfully tracked for at least 80\% of its life span is considered mostly tracked. This metric does not require that a player ID remains the same during the tracking.
If the target is tracked for less than 80\% and more than 20\% of its ground truth track, it's considered partially tracked (PT).
Otherwise, the target is mostly lost (ML).

Reported evaluation results for the synthetic dataset are averaged over four game episodes. 



\subsection{Results and discussion}
\label{sec:results}

\begin{figure}
\begin{center}
\includegraphics[width=1.0\linewidth]{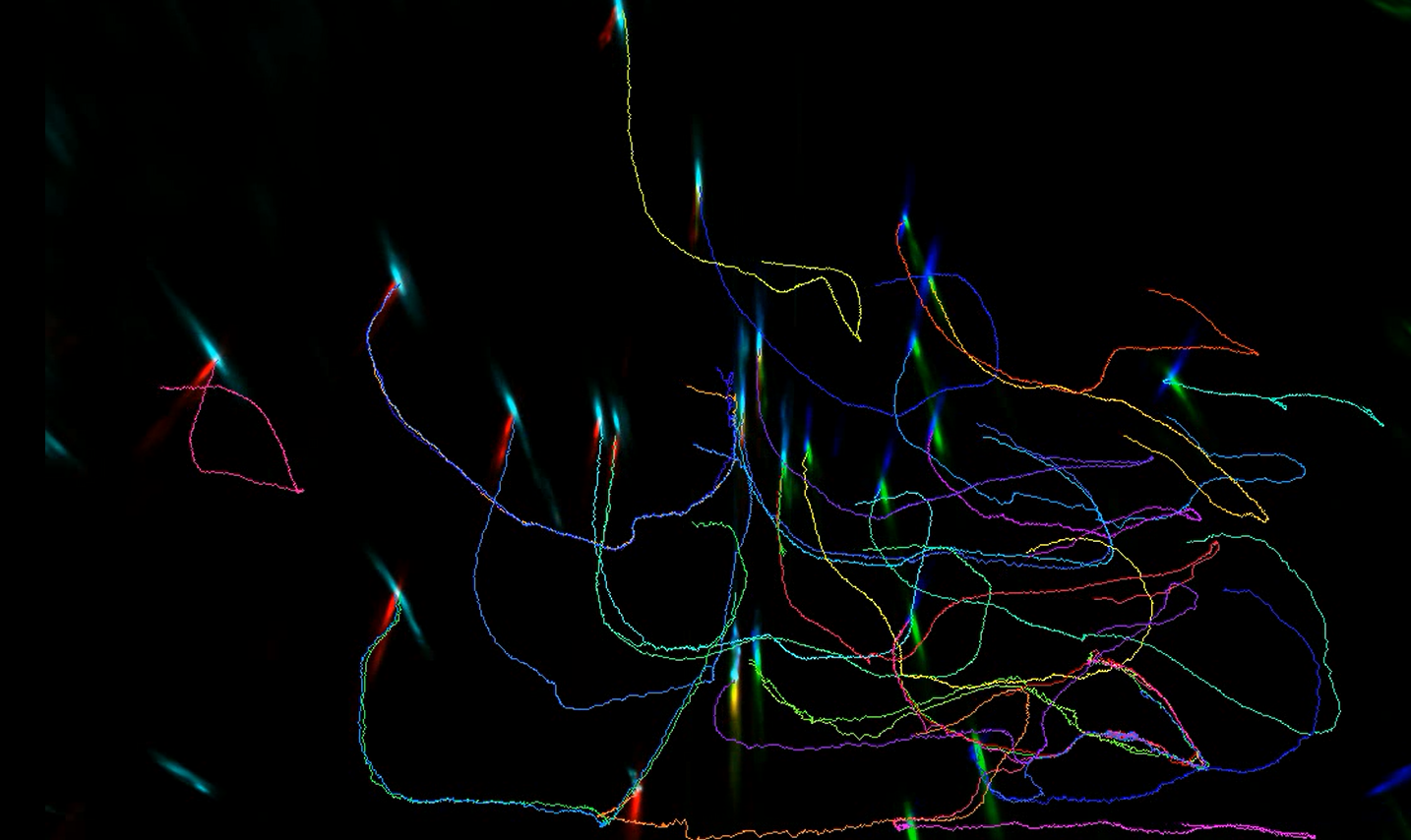}
\end{center}
\caption{Visualization of tracking results plot onto player detection heatmaps transferred to a bird's-eye view.
Detection heatmap values from each camera are color-coded for visualization purposes.}
\label{fig:tracking_results}
\end{figure}

Figure~\ref{fig:tracking_results} shows a visualization of tracking results superimposed onto detection heatmaps transformed to a bird's-eye view.
For visualization purposes, the detection heatmap from each camera is drawn with a different color.

\paragraph{Evaluation results.}
Table~\ref{jk:tab:results1} compares the performance of our proposed tracking method with the baseline solution.
The baseline tracker is a multiple target tracking method using particle filters~\cite{jinan2016particle}.
It's based on an aggregated bird's-eye view detection heatmap, constructed by transforming detection heatmaps from each camera to a ground plane view using a homography and summing them together.
Our method outperforms the baseline on all three test sets: a synthetic GRF set and real-world GT2 and GT3 sets.
MOTA metric is higher by 2-4\,p.p. Most importantly, the number of identity switches (ID) is significantly lower, by 50\% on average.
Without incorporating appearance cues, such as players' jersey numbers, our method significantly reduces identity switches compared to the baseline approach. This proves the validity and the potential of the proposed solution. 

\begin{table*}
\caption{Tracker evaluation results.
IDSW = number of identity switches, MT = mostly tracked, PT = partially tracked, ML = mostly lost. 
GRF is synthetic, and GT2, GT3 are real-world evaluation sets.}
\begin{center}
\begin{tabular}{l@{\quad}|r@{\quad}r@{\quad}r@{\quad}r@{\quad}r@{\quad}|r@{\quad}r@{\quad}r@{\quad}r@{\quad}r@{\quad}|r@{\quad}r@{\quad}r@{\quad}r@{\quad}r}
& \multicolumn{5}{c}{GRF (synthetic)} & \multicolumn{5}{c}{GT2 (real)} & \multicolumn{5}{c}{GT3 (real)}
\\
& \begin{tabular}{@{}c@{}}MOTA$\uparrow$ \end{tabular}
& \begin{tabular}{@{}c@{}}IDSW$\downarrow$ \end{tabular}
& \begin{tabular}{@{}c@{}}MT$\uparrow$ \end{tabular}
& \begin{tabular}{@{}c@{}}PT$\downarrow$ \end{tabular}
& \begin{tabular}{@{}c@{}}ML$\downarrow$ \end{tabular}
& \begin{tabular}{@{}c@{}}MOTA$\uparrow$ \end{tabular}
& \begin{tabular}{@{}c@{}}IDSW$\downarrow$ \end{tabular}
& \begin{tabular}{@{}c@{}}MT$\uparrow$ \end{tabular}
& \begin{tabular}{@{}c@{}}PT$\downarrow$ \end{tabular}
& \begin{tabular}{@{}c@{}}ML$\downarrow$ \end{tabular}
& \begin{tabular}{@{}c@{}}MOTA$\uparrow$ \end{tabular}
& \begin{tabular}{@{}c@{}}IDSW$\downarrow$ \end{tabular}
& \begin{tabular}{@{}c@{}}MT$\uparrow$ \end{tabular}
& \begin{tabular}{@{}c@{}}PT$\downarrow$ \end{tabular}
& \begin{tabular}{@{}c@{}}ML$\downarrow$ \end{tabular}
\\
[2pt]
\hline
\Tstrut
Baseline tracker & 0.823 & 13 & 20 & 2 & 0 & 0.925 & 30 & 25 & 0 & 0 & 0.828 & 41 & 24 & 1 & 0\\
\textbf{Tractor (ours)} & \textbf{0.874} & \textbf{7} & 20 & 2 & 0 & \textbf{0.949} & \textbf{11} & 25 & 0 & 0 & \textbf{0.867} & \textbf{23} & 24 & 1 & 0 \\
[2pt]
\hline
\end{tabular}
\end{center}
\label{jk:tab:results1}
\end{table*}

\paragraph{Ablation study.}

In this section, we analyze the impact of individual components of the proposed method on tracking performance.
In all experiments, we used a synthetic evaluation set (GRF).
Table~\ref{jk:tab:ablation1} shows the performance of reduced versions of our architecture versus the full model performance.
Disabling players' motion (players' movement trajectory) cues when regressing new player positions almost double the number of identity switches (ID), as shown in 'no player trajectory'. The tracking method is much more likely to confuse tracking object identities without knowing previous players' trajectories.
Switching off the message passing mechanism in the graph ('no message passing' row), where information is exchanged between neighborhood nodes (players), has a similar effect on the tracking performance. The number of identity switches increases from 7 to 12.
We can conclude that both components, player trajectory encoding using RNN and modeling neighborhood players interaction using GNN, are crucial for the good performance of the proposed method.

\begin{table}
\caption{
Performance of the full model compared to reduced versions.
Performance evaluated on the synthetic evaluation set.
MT = mostly tracked, PT = partially tracked, ML = mostly lost, IDSW = number of identity switches. 
}
\begin{center}
\begin{tabular}{l@{\quad}|r@{\quad}r@{\quad}r@{\quad}r@{\quad}r}
& \begin{tabular}{@{}c@{}}MOTA$\uparrow$ \end{tabular}
& \begin{tabular}{@{}c@{}}IDSW$\downarrow$ \end{tabular}
& \begin{tabular}{@{}c@{}}MT$\uparrow$ \end{tabular}
& \begin{tabular}{@{}c@{}}PT$\downarrow$ \end{tabular}
& \begin{tabular}{@{}c@{}}ML$\downarrow$ \end{tabular}
\\
[2pt]
\hline
\Tstrut
\textbf{Tractor (full model)} & \textbf{0.874} & \textbf{7} & 20 & 2 & 0 \\  
no player trajectory & 0.772 & 13 & 20 & 2 & 0 \\
no message passing &  0.776 & 12 & 20 & 2 & 0 \\ 
[2pt]
\hline
\end{tabular}
\end{center}
\label{jk:tab:ablation1}
\end{table}

Table~\ref{jk:tab:ablation2} the tracking performance for different choices of the detection heatmap embedding subnetwork.
We evaluated the following approaches: processing unaggregated crops from detection heatmaps  (each is a 4-channel, crop size by crop size tensor) using four-layer MLP with 1024, 512, 256, and 128 neurons in each layer (denoted as MLP);
aggregating crops from detection heatmaps using a convolutional layer with 1x1 kernel (this produces 1-channel tensor), followed by four-layer MLP with 1024, 512, 256, and 128 neurons in each layer (denoted as Mixed1 CNN+MLP);
summing crops from detection heatmaps across a channel dimension (this produces 1-channel tensor), followed by four-layer MLP with 1024, 512, 256, and 128 neurons in each layer (denoted as Mixed2 Sum+MLP);
using a 2D convolutional network with positional encoding using a CoordConv~\cite{liu2018intriguing} layer (denoted as CoordConv) and final global average pooling layer;
using a 2D convolutional network without the positional encoding and with the final global average pooling layer (denoted as Conv).
As expected, using a convolutional architecture without positional encoding to compute embedding of crops from input detection heatmaps yields worse results.
Convolutions are translation invariant and have difficulty extracting the spatial positions of detection heatmap maxima. This adversely affects the network's ability to predict the next position of each player. 
Better results are obtained by using a simple positional encoding with the CoordConv~\cite{liu2018intriguing} layer. 
CoordConv is used as a first network layer to encode spatial $x$, $y$ coordinates of each pixel in two additional channels, making the further processing coordinate-aware.
This improves the ability to localize detection heatmap maxima and improves the final tracking results.
The best results are achieved by concatenating detection heatmaps from multiple cameras (transformed to the bird's-eye view) using a convolution with 1x1 kernel and processing the resulting heatmap with a multi-layer perceptron (Mixed1: CNN+MLP).
This gives the highest tracking accuracy (MOTA) and the lowest number of identity switches.
Using multi-layer perceptron without aggregating detection heatmaps from multiple cameras yields worse results due to higher overfitting.

\begin{table}
\caption{
Performance of different architectures of the detection embedding subnetwork on the synthetic evaluation set.
MT = mostly tracked, PT = partially tracked, ML = mostly lost, IDSW = number of identity switches. 
}
\begin{center}
\begin{tabular}{l@{\quad}|r@{\quad}r@{\quad}r@{\quad}r@{\quad}r}
& \begin{tabular}{@{}c@{}}MOTA$\uparrow$ \end{tabular}
& \begin{tabular}{@{}c@{}}IDSW$\downarrow$ \end{tabular}
& \begin{tabular}{@{}c@{}}MT$\uparrow$ \end{tabular}
& \begin{tabular}{@{}c@{}}PT$\downarrow$ \end{tabular}
& \begin{tabular}{@{}c@{}}ML$\downarrow$ \end{tabular}
\\
[2pt]
\hline
\Tstrut
MLP & 0.796 & 12 & 20 & 2 & 0 \\
\textbf{Mixed1 (CNN+MLP)} &  \textbf{0.874} & \textbf{7} & 20 & 2 & 0 \\
Mixed2 (Sum+MLP) & 0.823 & 11 & 20 & 2 & 0 \\
CoordConv & 0.815 & 13 & 20 & 2 & 0 \\
Conv &  0.795 & 21 & 20 & 2 & 0 \\
[2pt]
\hline
\end{tabular}
\end{center}
\label{jk:tab:ablation2}
\end{table}

\section{Conclusion}

The paper presents an efficient multi-camera tracking method intended for tracking soccer players in long shot video recordings from multiple calibrated cameras.
The method achieves better accuracy and a significantly lower number of identity switches than a baseline approach, based on a particle filter.
Due to a large distance to the camera, visual cues, such as a jersey number, cannot be used.
Our method exploits other cues, such as a player movement trajectory and interaction between neighborhood players, to improve the tracking accuracy.
A promising future research direction is an integration of sparse visual cues, such as a jersey number which is readable in relatively few frames, into the tracking pipeline. This would allow further increasing the tracking accuracy.

\subsubsection*{Acknowledgments}
This study was prepared within realization of the Project co-funded by polish National Center of Research and Development, Ścieżka dla Mazowsza/2019.

{\small
\bibliographystyle{IEEEtran}
\bibliography{jkbib}
}

\end{document}